# Multi-Projector Color Structured-Light Vision

Changsoo Je [a,*], Kwang Hee Lee [b], Sang Wook Lee [b]

[a] Department of Electronic Engineering      [b] Department of Media Technology

Sogang University, 35 Baekbeom-ro, Mapo-gu, Seoul 121-742, Republic of Korea

[*]Corresponding author: vision@sogang.ac.kr      Office: +82-2-711-8916     Fax.: +82-2-706-4216

**Abstract**

Research interest in rapid structured-light imaging has grown increasingly for the modeling of moving objects, and a number of methods have been suggested for the range capture in a single video frame. The imaging area of a 3D object using a single projector is restricted since the structured light is projected only onto a limited area of the object surface. Employing additional projectors to broaden the imaging area is a challenging problem since simultaneous projection of multiple patterns results in their superposition in the light-intersected areas and the recognition of original patterns is by no means trivial. This paper presents a novel method of multi-projector color structured-light vision based on projector-camera triangulation. By analyzing the behavior of superposed-light colors in a chromaticity domain, we show that the original light colors cannot be properly extracted by the conventional direct estimation. We disambiguate multiple projectors by multiplexing the orientations of projector patterns so that the superposed patterns can be separated by explicit derivative computations. Experimental studies are carried out to demonstrate the validity of the presented method. The proposed method increases the efficiency of range acquisition compared to conventional active stereo using multiple projectors.

## 1. Introduction

Depth imaging is a critical step in object modeling, and structured light-based range sensing is probably the most accurate and reliable way of obtaining depth data. Each depth image contains part of a 3D object model, and several depth images should be combined to represent a 3D object. Modeling a moving object requires depth capture from multiple views simultaneously in realtime, but structured light is not easy to use since multiple projections of active lights generally result in the superposition of light patterns on some parts of the object surfaces. The objective of the work described in this paper is to develop a method for projecting multiple light patterns in one video frame (one shot) and recognizing those to reconstruct the shapes of different but overlapping object regions.

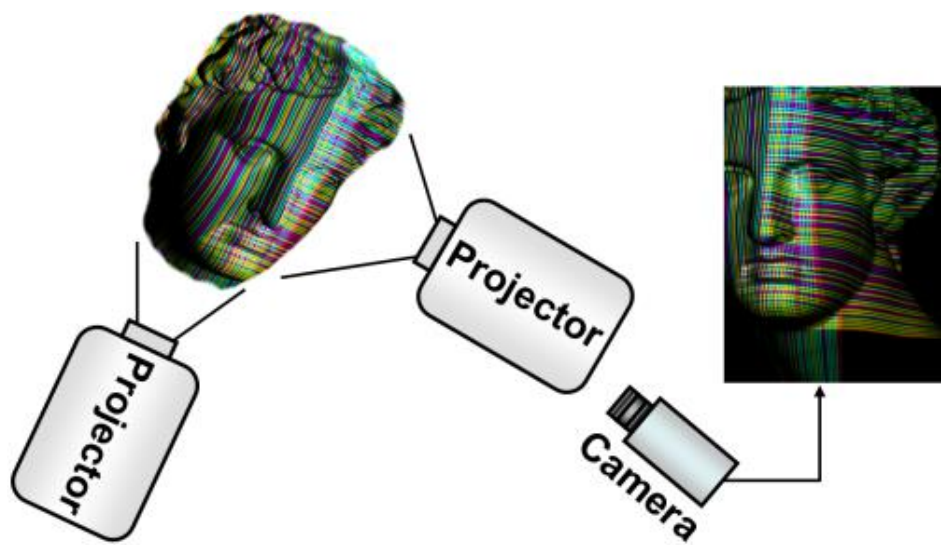

Figure 1. Multi-projector structured-light range imaging. Two projectors project the light patterns onto an object, and a camera observes the scene. The light patterns are superposed in the intersection of two projections.

Structured light has been an active area of research for decades. The most basic form of depth sensing is the swept stripe scanner on which much research has been carried out [1, 2]. Its main disadvantage is the large number of images required to be taken as a plane of light sweeps across an object surface. Sato and Inokuchi developed a set of hierarchical black-and-white stripe patterns to reduce the number of images to $\log_2 N$ where $N$ is the total number of stripes [3]. For more drastic reduction, a variety of special black-and-white, grey-level, or color patterns have been presented: color-stripe patterns [10, 4, 33], hybrid pattern [24], grey-level pattern [5], gray-level ramp [6], rainbow pattern [7], black-and-white boundary codes [21], 2D stripe patterns (or grid patterns) [26, 28, 29], M-array pattern [27], color sinusoidal patterns [8, 22], color dot pattern [9], de Bruijn sequence patterns [12, 13], and permutation patterns [11, 31]. Some researchers have focused on scene properties [23], and scattering environment [25].

With a single projector, a structured-light pattern is projected only onto a partial area of a 3D object due to self shadows (surface orientation being more than $\pi/2$ away from the lighting direction) and cast shadows (light occlusion by other parts). The only realistic way of increasing the imaging area is to use light sources from multiple directions. For modeling a static object, structured-light ranging is performed sequentially from multiple directions, and the acquired multiple depth images are aligned and merged using sensor/light calibration or range data registration technique. For modeling a dynamic object, on the other hand, simultaneous multi-view ranging is required but multiple structured-light patterns from different directions interfere with each other in the intersection of light projections as shown in Figure 1. This overlap of light patterns is unavoidable for general 3D object geometry unless an undesirable gap between the projected areas is made deliberately. Seamless modeling of a dynamic object can be made highly practicable if we can extract the original light patterns superposed in the intersections. This motivates our work presented in this paper.

Table 1 shows taxonomy of active vision. Conventional active stereo vision methods based on camera-camera triangulation are in general capable of real-time depth imaging. They add structured light to create uniquely identifiable features for dense stereo correspondence. Light patterns include stripes [14, 15, 16], circular color spots [9], and IR random dots [17]. Recently some researchers have begun using multiple projectors for imaging wider surface areas, and they obtained the shape of a moving object from a small number of frames [18] or from a single frame [17]. Zhang *et al*. use spacetime matching of active stereo correspondences by employing time-varying grayscale random stripe patterns [16, 18], and Ypsilos *et al*. use a random dot IR pattern for stereo matching [17].

When multiple structured-light patterns are simultaneously projected onto an object to increase the imaging area seamlessly, the patterns are almost always superposed in the intersections of projections (see Figure 1). This may not cause a serious problem for conventional active stereo vision (ASV) based on camera-camera triangulation. However, in conventional structured-light vision (SLV) based on projector-camera triangulation, extracting the original light patterns is the core problem, and this has hindered multi-projector SLV. In this paper, we present a method of multi-projector color structured-light range imaging based on projector-camera triangulation. By analyzing the behavior of the superposed-light colors in a chromaticity domain, we show that the original light colors cannot be properly extracted based on the conventional direct estimation. We disambiguate multiple projectors by multiplexing the orientations of projector patterns so that the superposed patterns can be separated by explicit derivative computations. Experimental studies are carried out to demonstrate the validity of the presented method.

Table 1
Active Vision Taxonomy

|  | Camera vs. Camera | Projector vs. Camera |
|---|---|---|
| *Single pattern* | Conventional ASV : Chen *et al.* [15] | Existing SLV: Boyer and Kak [10] Caspi *et al.* [4] Hall-Holt and Rusinkiewicz [21] Zhang et al. [12] Je et al. [11, 31] |
| *Multiple patterns* | Conventional ASV : Zhang *et al.* [16, 18] Ypsilos *et al.* [17] | Multi-projector SLV: Our approach [32] Furukawa *et al.* [30] |

This paper is mostly based on the sixth chapter of Je's PhD thesis presented early in 2008 [32], and provides improved analyses on color behavior in multi-projector, separation of superposed patterns, and estimation of stripe encoding directions, as well as a scheme of range merging and diverse comparisons of related methods. In 2010, Furukawa *et al.* independently presented a closely related method [30]. Nevertheless, our approach still has strong advantages compared to the method of Furukawa *et al.* [30] although our method has been originally presented much earlier. We will present the comparison of the two methods in Section 5.

The rest of this paper is organized as follows. Section 2 analyzes the color behavior in multi-projector illumination, Section 3 presents the orientation multiplexing in multi-projector, and the experiments are shown in Section 4. Section 5 discusses on the related shape-measuring methodologies, and Section 6 concludes this paper.

**2. Color behavior in multi-projector illumination**

In range imaging based on SLV, the range of object surface can be reliably obtained only for the area in which the intensity profile is sufficiently stronger than the random noise level and in which the viewing space of a camera and the lighting space of a projector are intersected. In single-projector SLV, images frequently have dark areas due to shadows (self shadows and/or cast shadows), shading, and camera settings selected to avoid extreme saturation of highlight areas (e.g., specularity).

Reliable range imaging for the whole area in an image is almost impossible in a typical-complicate situation using a single projector for the following reasons. (1) The spaces of viewing and lighting are usually different in that the projector and camera components cannot be at the exactly same position. (2) The triangulation angle should be sufficiently large for accurate measurement. For example, the baseline of a Kinect [34] depth sensor is about 7.96 cm that is smaller than a baseline in a typical projector-camera configuration. If the distance from a Kinect depth sensor to an object is bigger than 1 m, the triangulation angle is smaller than 4.6 degrees. As a result, a Kinect depth sensor acquires only coarse shapes of 1 m or more distant objects. In general, the triangulation angle should be substantially bigger than 5 degrees for acquiring satisfactory shape of an object, resulting in smaller overlap of lighting and sensing spaces. To make matters worse, a larger triangulation angle for the accuracy makes the effects of shadows and shading bigger. Recently, systems employing multiple cameras have increased in popularity, requiring multiple light sources more often. However, in SLV, employing multiple projectors has been prohibited since superposition of multiple patterns makes the original patterns ambiguous. This means multi-projector makes the resulted state far from the original objective of structured light.

Now we discuss the behavior of colors of the superposed patterns in multi-projector structured light. The reflected intensity of a wavelength ($\lambda$) on a Lambertian surface point can be given by the equation,

$$I(\lambda) = g_\theta S(\lambda) E(\lambda) \tag{1}$$

where $S$, $g_\theta$, and $E$ are the surface reflectance, geometric shading factor, and the illumination, respectively, and the wavelength $\lambda$ represents color.

When light patterns from two projectors are superposed with ambient illumination on a surface, the image irradiance can be modified by the equation,

$$I(\lambda) = S(\lambda)[g_1 E_1(\lambda) + g_2 E_2(\lambda) + g_A A(\lambda)] \tag{2}$$

where $g_1$, $g_2$, $g_A$, $E_1$, $E_2$, and $A$ are the geometric shading factors and the illumination from the projectors 1 and 2, and the ambient light, respectively. From the above equation, the number of unknowns (the illumination intensities from both projectors) for a known (the image intensity) is two at least even if the reflectance, shading, and ambient illumination are ignored by the assumption that the spatial variations of them in a camera image are not so severe usually, compared to those of structured-light patterns.

Let us assume projectors and camera(s) have three color-channels (R, G and B). The pixel value corresponding to a scene point on the object surface illuminated by multiple projectors is a combination of three scalar values of red (R), green (G) and blue (B). However, the number of unknowns includes at least three (RGB) times of the number of light sources illuminating the scene area even under the assumption that the reflective properties are perfectly diffuse and spectrally moderate. For example, in the case of two projectors the number of unknowns is six. Therefore, it is almost impossible to acquire the colors of multiple superposed patterns from the image color unless other exceptional information is provided.

Figure 2(a) and 2(b) depict the color distribution where the three primary colors (RGB) are projected separately onto an achromatic surface and a moderate-colored surface by a single projector, respectively. Figure 2(c) shows the color distribution where the three colors are projected by two projectors and the colors are superposed. In Figure 2(c), it is shown that when two different colors (from the two different projectors) are superposed, and the intermediate colors are made.

Figure 3(a) and 3(b) depict the color distribution where concatenated RGB stripes are projected onto an achromatic surface and a moderate-colored surface, by a single projector, respectively. Figure 3(c) shows the color distribution where the RGB stripes are projected onto a moderate-colored surface by two projectors. In Figure 3(a) and 3(b), it is shown that there comes the intermediate colors between the two consecutive color stripes. Figure 3(c) depicts the intermediate colors are superposed, and an arbitrary combination of the three primary colors is produced.

In Figure 2(a), 2(b), 2(c), 3(a) and 3(b), the colors can be easily separated and the stripes can be properly segmented by a suitable method like *hue thresholding* [11, 31]. However, in Figure 3(c), it cannot be easily known which two colors made a color. In addition, in Figure 2(c) and 3(c), even if the two colors can be correctly estimated, it cannot be easily known which projector produced each of the estimated colors.

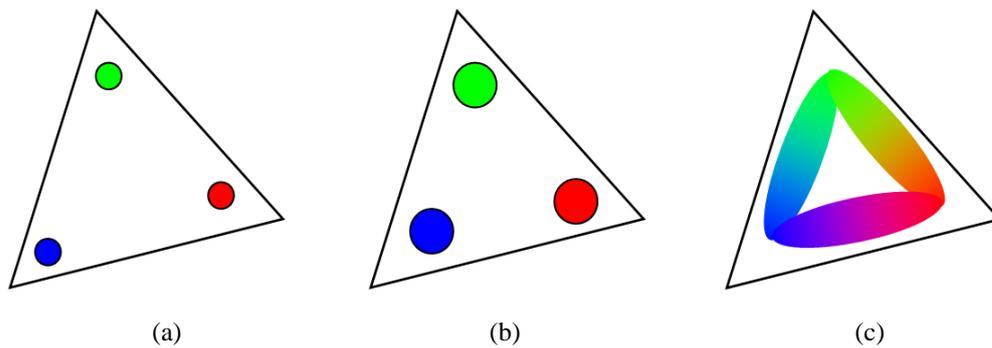

Figure 2. Chromaticity diagram: (a) RGB projection onto an achromatic surface by a single projector, (b) RGB projection onto a moderate-colored surface by a single projector, and (c) RGB projection onto a colored surface by two projectors.

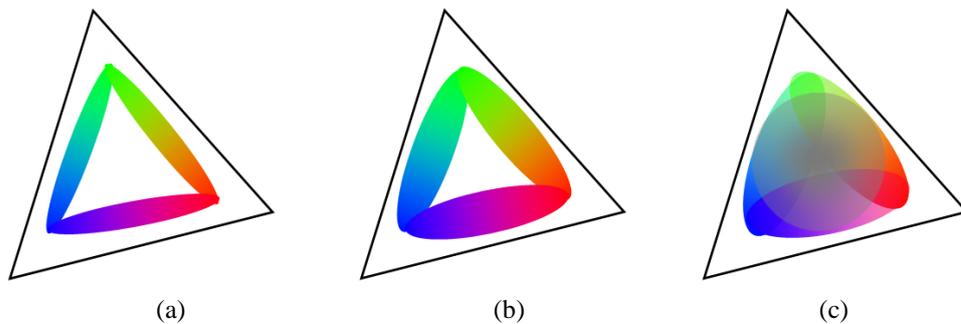

Figure 3. Chromaticity diagram: (a) RGB stripes projection onto an achromatic surface by a single projector, (b) RGB stripes projection onto a moderate-colored surface by a single projector, and (c) RGB stripes projection onto a colored surface by two projectors.

## 3. Orientation Multiplexing in Multi-projector

As discussed in Section 2, direct estimation of the original color in multi-projector is not possible. We add another cue, say, *orientation cue* of the projection patterns to distinguish multiple patterns. We disambiguate the patterns of different projectors by projecting multi-projector patterns in dissimilar orientations. As we use different colors in two consecutive stripes to distinguish the stripes in a color-stripe pattern, we use different orientations in simultaneous projections by multi-projector to distinguish the projections.

### 3.1. Derivative Analysis for Orientation Demultiplexing

We now describe derivative analysis based on two discrete color-stripe patterns projected in substantially different orientations by two different projectors, but it can be extended to more than two projectors.

The image irradiance at a camera is given by Equation (2). The exact extraction of the irradiance components from $E_1$ and $E_2$ is an underconstrained problem; hence our method seeks to find the spatial derivative terms of $E_1$ and $E_2$ separately using the following constraints described below. In other words, we are not interested in estimating the actual image intensities and colors from $E_1$ and $E_2$ but in recovering the color codes encoded in $E_1$ and $E_2$ from their derivative terms.

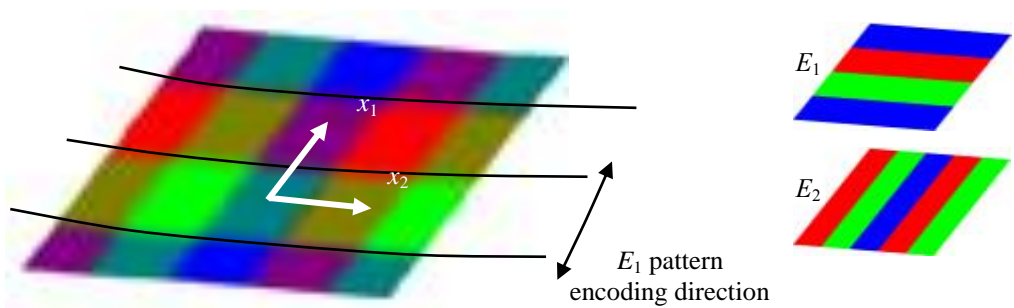

Figure 4.  Local variables and directions. $x_1$ and $x_2$ are the encoding directions of $E_1$ and $E_2$, respectively.

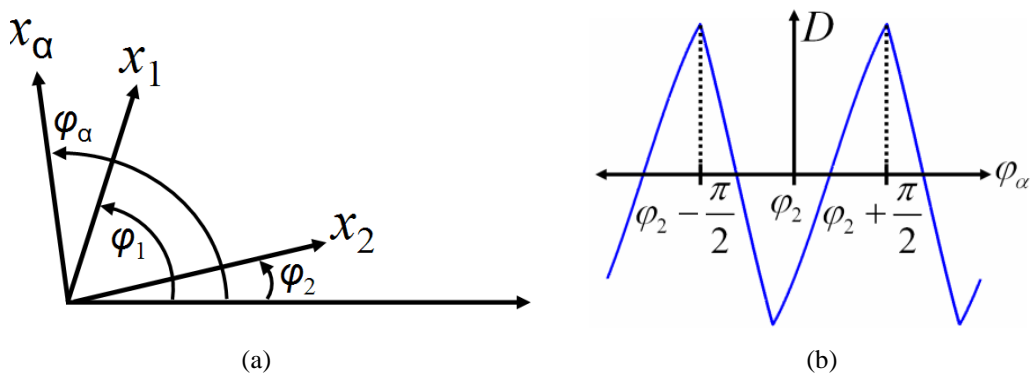

(a)                                                  (b)

Figure 5.  (a) The encoding directions of the two patterns $E_1$ and $E_2$, and the differentiation direction $x_\alpha$. (b) The plot of $D$ in Equation (7) for $\varphi_1 - \varphi_2 = 70°$.

Since a light pattern is one-dimensionally encoded, its derivatives across the stripes should not be much affected by the presence of another pattern as far as their directions are maximally different. If it is assumed that the two patterns are exactly perpendicular, for instance, the partial derivative with respect to the spatial variable, $x_1$ of the encoding direction (say, $\hat{x}_1$) of $E_1$ (see Figure 4) is:

$$\frac{\partial I(\lambda)}{\partial x_1} = g_1 S(\lambda) \frac{\partial E_1(\lambda)}{\partial x_1}. \tag{3}$$

In the above equation, it is also assumed that the local variation of image irradiance due to geometric shading, reflectance and ambient light is negligible compared to that of $E_1$. Since subpatterns based on the spatial derivatives of color across the stripes are as unique as those based on stripe colors, they can be used for stripe (boundary) identification and for triangulation.

As a matter of fact, the stripe directions on an object surface are not mostly made mutually perpendicular due to the projector perspectivity and local surface geometry. When two light patterns are superposed in arbitrary orientations, the partial derivative with respect to the spatial variable, $x_\alpha$ of an arbitrary direction ($\hat{x}_\alpha$) (see Figure 5 (a)) is given by:

$$\frac{\partial I(\lambda)}{\partial x_\alpha} = S(\lambda)[g_1 \frac{\partial E_1(\lambda)}{\partial x_1} \frac{\partial x_1}{\partial x_\alpha} + g_2 \frac{\partial E_2(\lambda)}{\partial x_2} \frac{\partial x_2}{\partial x_\alpha}], \tag{4}$$

where $x_2$ is the spatial variable of the encoding direction of $E_2$. Since $\partial x_1/\partial x_\alpha = \cos(\varphi_\alpha - \varphi_1)$ and $\partial x_2/\partial x_\alpha = \cos(\varphi_\alpha - \varphi_2)$ as shown in Figure 5(a), the partial derivatives of $E_1$ and $E_2$ in Equation (4) are respectively scaled by the cosines of the angle between $x_1$ and $x_\alpha$ and that between $x_2$ and $x_\alpha$, *i.e.*, the scale factors vary depending on the selection of $x_\alpha$. It is now obvious that the influence of the light pattern $E_2$ can be minimized by choosing $x_\alpha$ perpendicular to $x_2$. Notice that this choice of $x_\alpha$ also maximizes the difference of the absolute values of the two derivative scaling factors in Equation (4) (Let $D$ be the difference),

$$D = \left|\frac{\partial x_1}{\partial x_\alpha}\right| - \left|\frac{\partial x_2}{\partial x_\alpha}\right| = |\cos(\varphi_\alpha - \varphi_1)| - |\cos(\varphi_\alpha - \varphi_2)| \tag{5}$$

since it has the maximums at $\varphi_\alpha = \varphi_2 \pm \pi/2$. Figure 5(b) shows the plot of $D$ for $\varphi_1 - \varphi_2 = 70°$.

As far as we have good estimations of $x_1$ and $x_\alpha$, we can extract the derivative information on the projected patterns by taking the derivatives as follows:

$$\frac{\partial I(\lambda)}{\partial y_2} = g_1 S(\lambda) \frac{\partial E_1(\lambda)}{\partial x_1} \frac{\partial x_1}{\partial y_2}, \tag{6}$$

$$\frac{\partial I(\lambda)}{\partial y_1} = g_2 S(\lambda) \frac{\partial E_2(\lambda)}{\partial x_2} \frac{\partial x_2}{\partial y_1}, \tag{7}$$

where $y_1$ and $y_2$ are perpendicular to $x_1$ and $x_2$, respectively. We can obtain the shape independently based on the derivative of the projected pattern from Equations (6) and (7) (the range estimation process is described in Section 4.2). When $N$ multiple patterns are superposed on a surface in $N$ different orientations, we can estimate the term only from one pattern by successively eliminating the influence of ($N-1$) patterns. For example, triple-pattern case is as follows.

Let the image irradiance from three superposed patterns as:

$$I(\lambda) = S(\lambda)[g_1 E_1(\lambda) + g_2 E_2(\lambda) + g_3 E_3(\lambda) + g_A A(\lambda)]. \tag{8}$$

We assume again that the local variation of image irradiance due to geometric shading, reflectance and ambient light is negligible compared to that of $E_1$, $E_2$, and $E_3$. By differentiating this according to the spatial variable of the direction perpendicular to the encoding direction of the third pattern, the pattern term can be eliminated:

$$\frac{\partial I(\lambda)}{\partial y_3} = S(\lambda)[g_1 \frac{\partial E_1(\lambda)}{\partial x_1} \frac{\partial x_1}{\partial y_3} + g_2 \frac{\partial E_2(\lambda)}{\partial x_2} \frac{\partial x_2}{\partial y_3}]. \tag{9}$$

Then, the second pattern can be eliminated by differentiating the above equation according to the direction perpendicular to the encoding direction of the pattern:

$$\frac{\partial^2 I(\lambda)}{\partial y_2 \partial y_3} = g_1 S(\lambda) \frac{\partial^2 E_1(\lambda)}{\partial^2 x_1} \frac{\partial^2 x_1}{\partial y_2 \partial y_3}. \tag{10}$$

The only difference from the double-pattern case is that the order of the final derivative is two. Hence, we should decode the final profile according to the sequence of the second-order derivative of the original pattern. By the same way, the second and third patterns can be extracted and decoded. Although this pattern extraction method can be applied for general *N*-pattern case, in a real situation (limited dynamic range and resolution) stable results cannot be drawn for the superposition of so many patterns. In our work, we have implemented the method for the superposition of double- and triple-pattern. It is known that differentiations usually raise sensitivities to high-frequency noise. In case of triple-pattern, differentiation is needed twice for extracting each single pattern. However, successful separation of the stripe patterns can be achieved by using highly distinguishable stripe colors compared to noise uncertainties. Otherwise, slight Gaussian smoothing (with a small radius, e.g., 0.5 pixels) can be used for decreasing the effect of high-frequency noise.

In the pattern extraction, we need to determine the stripe-encoding direction to differentiate the images. The encoding directions can be assumed globally constant if the object geometry is not relatively steep. Otherwise, the stripe directions should be locally estimated. In the following subsection, we describe a method for computing stripe-encoding directions $\hat{x}_1$ and $\hat{x}_2$ locally.

### 3.2. Estimation of Stripe Encoding Directions

Successful extraction of structured light patterns depends on the estimation of their stripe encoding directions. We take rather a generic approach to this based on gradient histogram.

The gradients of individual RGB color-channels that can exist in a region of two color-stripe patterns are illustrated in Figure 6(a), in which for the sake of simplicity the two color-stripe patterns are assumed perpendicular. Figure 6(c) shows typical directions of gradients in a case of non-perpendicular patterns. In addition to the gradients along the stripe-encoding directions $\hat{x}_1$ and $\hat{x}_2$, there can be fringe gradients (along roughly $\hat{x}_{12}$ and $\hat{y}_{12}$) across the corners made by the two stripe patterns. Fortunately, the population of the fringe gradients is usually substantially smaller than that of the gradients along $\hat{x}_1$ and $\hat{x}_2$, since the corner area (around the intersection of boundaries) is definitely smaller than the single-boundary area in an image (see the split RGB-channel images in Figure 6(b)).

Before a gradient histogram is constructed, the gradient angles are multiplied by two to make the gradients in the opposite directions have the same polarity. Then, a doubly angled histogram will have the same number of dominant lobes as the number of superposed patterns in a region. Figure 6(d) illustrates a doubly angled gradient histogram that has two dominant lobes (bimodal) corresponding to the two superposed patterns in a region. When an area includes a single pattern the histogram will have a single dominant lobe (unimodal). Figure 7 shows how the gradient vectors are distributed in a pattern-superposed region. (a) shows a synthesized image of superposed two color-stripe patterns containing moderate noise (1.3% Gaussian noise), and (b) depicts the double-angled gradient-vector distribution of (a). As shown in Figure 7(b) the two dominant directions of gradient vectors can be easily estimated. Compare Figure 7(b) with the illustrative gradient histogram in Figure 6(d).

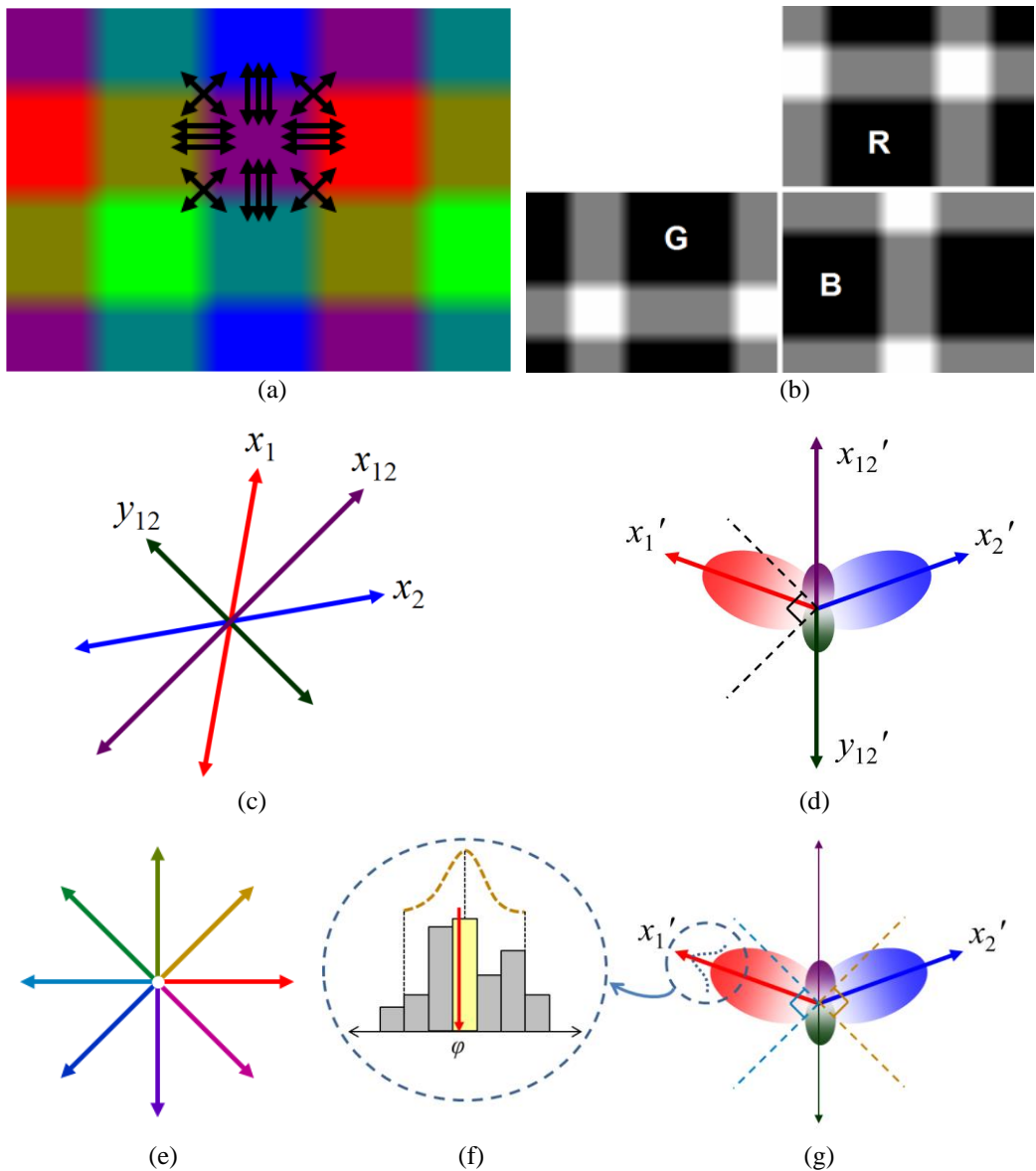

Figure 6. (a) Gradients which can occur where two color-stripe patterns are superposed in different orientations. (b) The split RGB-channel gray images of (a). (c) Classification of gradients. The direction of $x_{12}$ is equal to the mean direction of $x_1$ and $x_2$, and the direction of $y_{12}$ is perpendicular to that of $x_{12}$. (d) Gradient histogram. The angle of each gradient is scaled up by two. (e) The eight angular locations for an angle window are illustrated by distinct arrows in unique colors. (f) Gaussian-weighted fitting. The bright yellow rectangle represents the initial peak, and the red arrow indicates the fitting result. (g) The estimated encoding directions and the instances of the angle window.

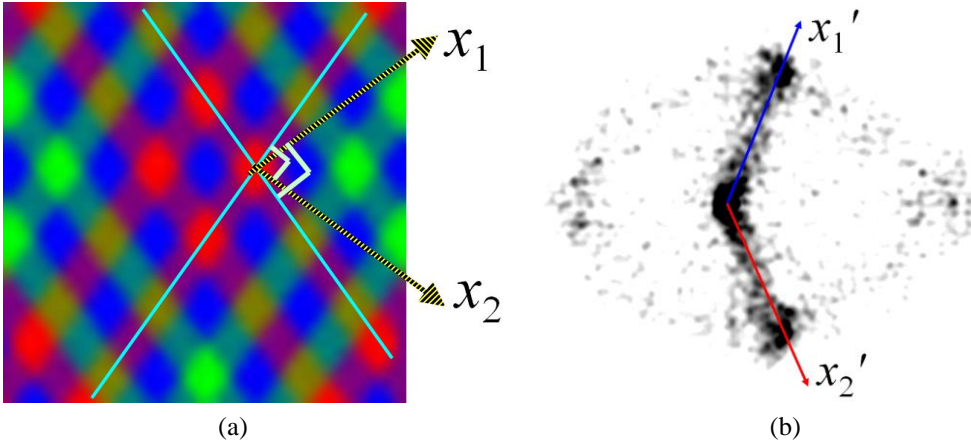

(a) (b)

Figure 7. Gradient-vector distribution of a synthesized image. (a) A synthesized image of superposed two color-stripe patterns containing moderate noise. (b) The double-angled gradient-vector distribution of (a). The darker area has the higher population. The two dominant directions of gradient-vectors are the two directions of $x_1$ and $x_2$. The direction of the primed is the double-angled direction of the unprimed.

To find the lobes along the $\hat{x}_1$ and $\hat{x}_2$ directions locally, we run a spatial window of 7×7 for localization. We also run an angle window in the histogram with the width of $\pi/2$ and scan the histogram in $\pi/4$ steps as shown in Figure 6(e). In this angle window, only one major lobe is expected to be present. Figure 6(d) shows an angle window centered at $\pi$. Once a peak is detected by finding the maximum value in an angle window, we locally fit a Gaussian function using the values around the peak to find the major direction in a manner similar to that in [19] as shown in Figure 6(f) and (g). After detecting lobes in the eight angle windows, one or two major directions (or also three directions in a triple-pattern case) are selected for $\hat{x}_1$ and $\hat{x}_2$ as shown in Figure 6(g).

Different size of window can be selected for localization of stripe encoding directions. The width and height of a window should be bigger than stripe width in a camera image. To substantially localize the encoding direction, we select 7 pixels for the width and height of window, which is near twice stripe width in our camera images. However, bigger window size can be selected when spatial variation of stripe encoding directions is not much in a camera image.

## 4. Experiments

We have tested our multi-projector method for various real objects. For capturing real scene images, we used Epson EMP-7700 1024×768 color LCD projectors and Sony XC-003 3-

CCD 640×480 color cameras. A small number of the color-stripe permutation patterns with 192 unique codes are projected on the object scene with mutually distant orientations by the projectors. The mutual angles of the orientations are roughly 90 degrees and 60 degrees in double-pattern and triple-pattern case, respectively. For imaging the scene, one or two cameras are employed according to the target area. The captured images are processed for obtaining the ranges, and for a human face and a Venus face, the ranges are meshed and rendered.

In the following subsection, we describe the used color-stripe pattern.

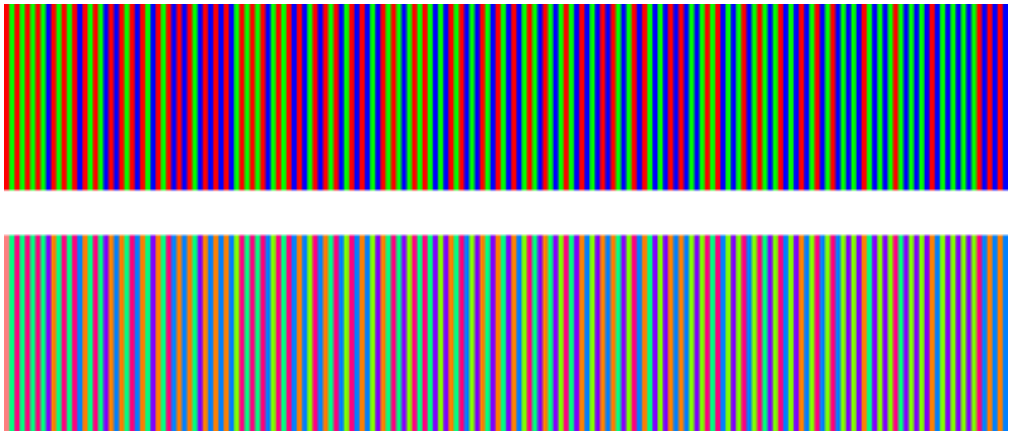

Figure 8. Color-stripe permutation pattern for multi-projector structured-light range imaging. (Top) the color-stripe pattern is generated based on the permutation of three primary colors (R, G, and B) with length seven. (Bottom) image representation for the horizontal derivative of the pattern (255 is added to the original derivative, and the result is divided by 2 to make a proper image intensity).

### 4.1. Color-Stripe Permutation Pattern

We employ the one-shot imaging pattern described in [11, 31, 35] for light projection, but use its spatial derivatives for the separation of superposed stripe patterns and for stripe identification. Only the three primary colors, red (R), green (G) and blue (B), are used for generating a pattern since the use of the most distant three colors in a chromaticity space facilitates accurate color stripe segmentation according to the analysis shown in [11, 31]. With the number of stripes in a subpattern, $k=7$ and the number of colors, $N=3$, 192 uniquely identifiable subpatterns can be generated as shown in the top image of Figure 8. The subpatterns based on the color derivatives across the stripes, $\pm RG$, $\pm GB$, $\pm BR$, have the same level of uniqueness as those based on stripe colors, and the color derivatives shown in the bottom image of Figure 8 are also maximized by using the three primary RGB colors. As discussed in [31], the color-stripe pattern is robust to surface discontinuity since sufficiently many stripes can be projected onto a very narrow continuous region.

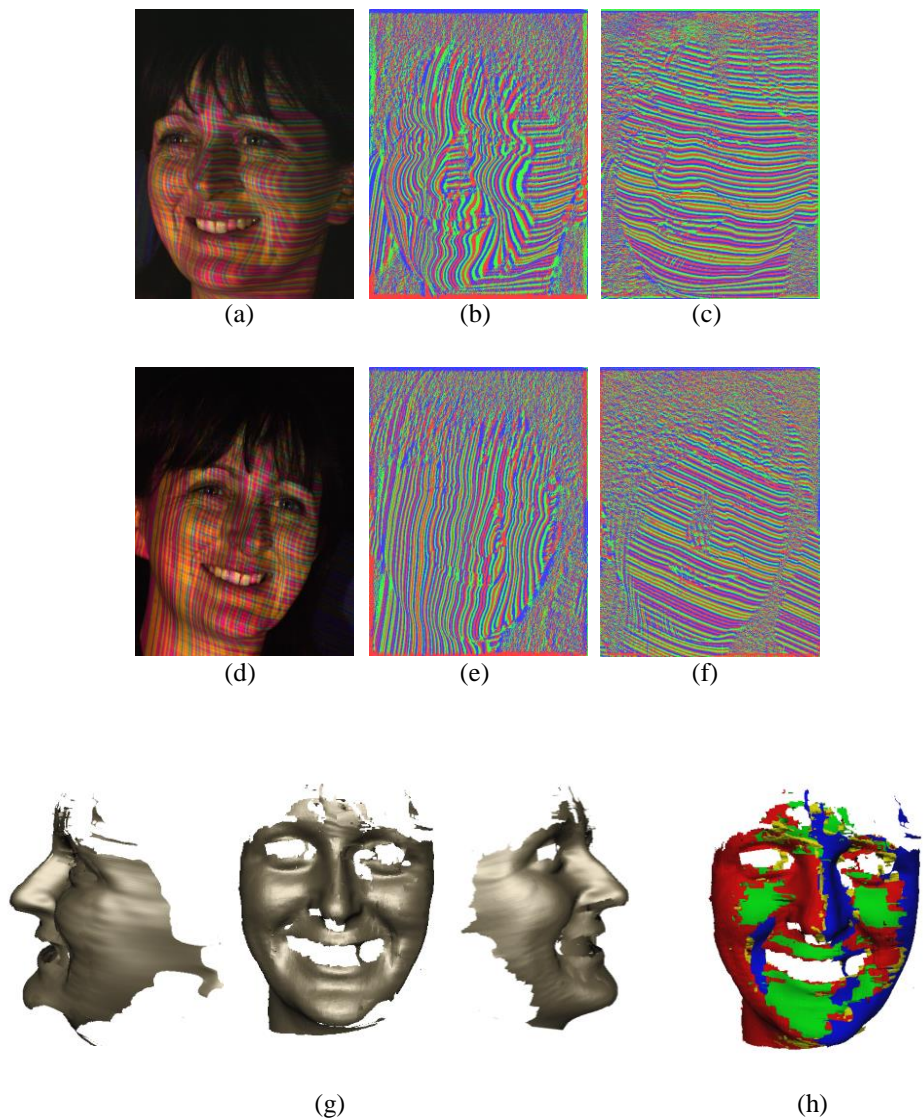

Figure 9. Experimental results of a human face. (a) Right view of the color-striped human face, and (b,c) stripe-segmentation from the gradient image of (a). (d) Left view of the subject, and (e,f) stripe-segmentation from the gradient image of (d). (g) Merged and rendered result in three different views, and (h) the rendered front view of the four individual meshed results from the four stripe-segmented images (b), (c), (e) and (f) in different colors.

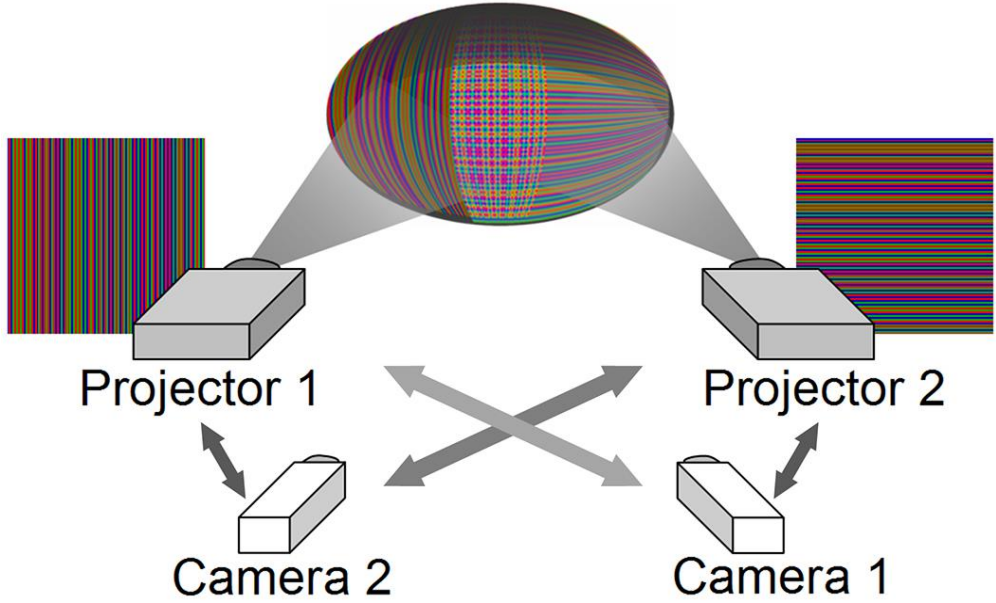

Figure 10. The arrangement of projectors and cameras. Two projectors project light patterns onto an object from the different angles and two cameras observe the scene with the different viewpoints. In this configuration, the proposed SLV (Structured-Light Vision) utilizes the four correspondences (projector1-camera1, projector1-camera2, projector2-camera1, and projector2-camera2), and obtains the four range images from a single video-frame.

### 4.2. Range Estimation Process

The process of range estimation from an image is as follows. The superposed light patterns have to be separated by differentiating the image through the estimated encoding directions for range imaging from each projector as described in Section 3. After separating derivative images, we segment each stripes based on each pixel color of the separated images by classifying the derivative colors into the six classes ($\pm$RG, $\pm$GB, $\pm$BR) and then classifying the colors into the three classes, R={+BR, −RG}, G={+RG, −GB}, and B={+GB, −BR}. Now the color-stripe permutations are decoded by matching the designed codes to the color sequences of the stripe-segmented image. To diminish decoding errors, we perform *multi-layer decoding* as described in [31]. An appropriate geometric calibration for SLV [20] is done for obtaining range data from the decoded stripe information. In data merging stage, we merge the ranges from the separated patterns for each image first, and then merge the results from the multiple cameras when multiple cameras are used.

**Noise effect reduction**: In our experimental system, the measured standard deviations of spacetime (50 frames of 7×7 window) noise for red, green and blue channels are 1.8138, 1.2923 and 1.6745 (in units of gray levels), respectively, and they are smaller than those in

[4]. The noise was measured by the setting that the intensity levels are close to 128 in the camera image, to avoid the clipping error. One method of discounting the effect of high-frequency noise is Gaussian smoothing.

### 4.3. Results

**Dynamic object**: Figure 9 shows the experimental results for a human face. Two cameras and two projectors are employed to capture a broad area of the face (see Figure 10). Four depth images are obtained from the four triangulation instances between the two projectors and two cameras: Projector 1 vs. Camera 1, Projector 1 vs. Camera 2, Projector 2 vs. Camera 1, and Projector 2 vs. Camera 2. On the surfaces where multiple light patterns are superposed, four different range data are obtained and their median is taken for merging.

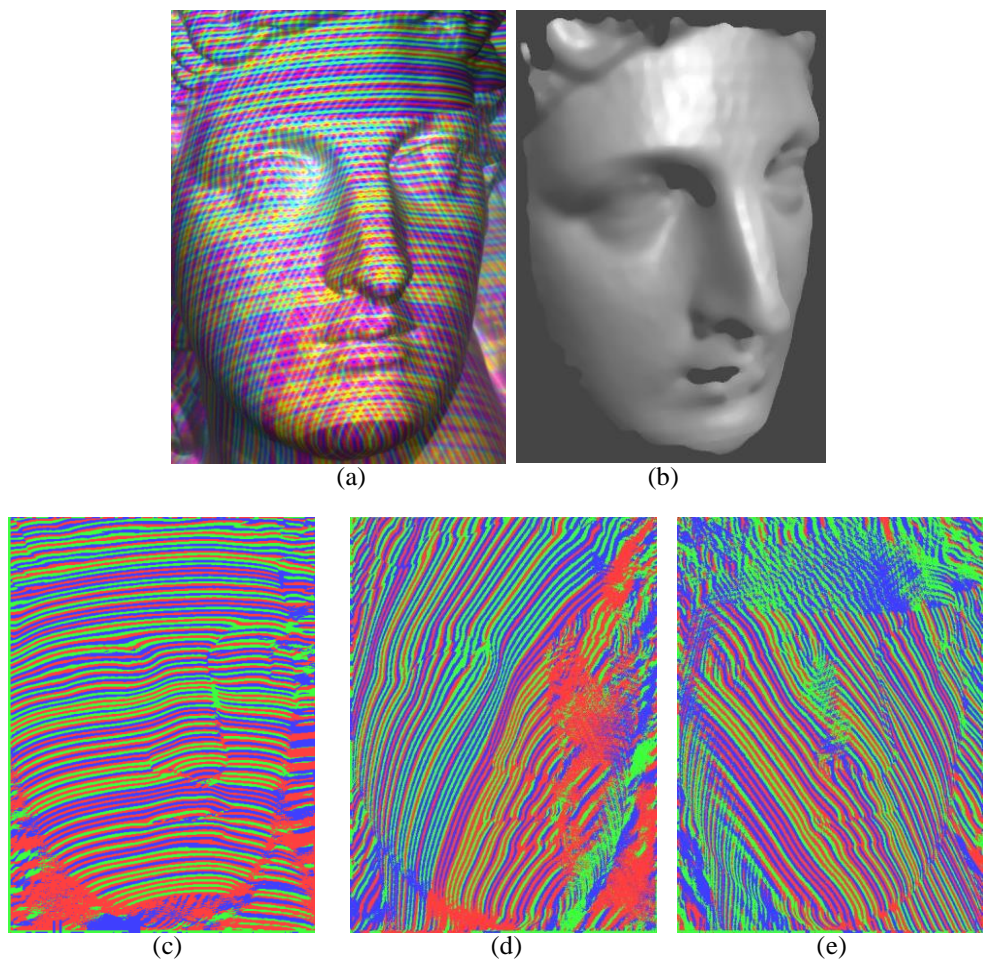

Figure 11. (a) A scene of Venus face illuminated by three patterns, (c,d,e) the three separated stripe-segmented images, and (b) rendered view of the merged range obtained from the three stripe-segmented images.

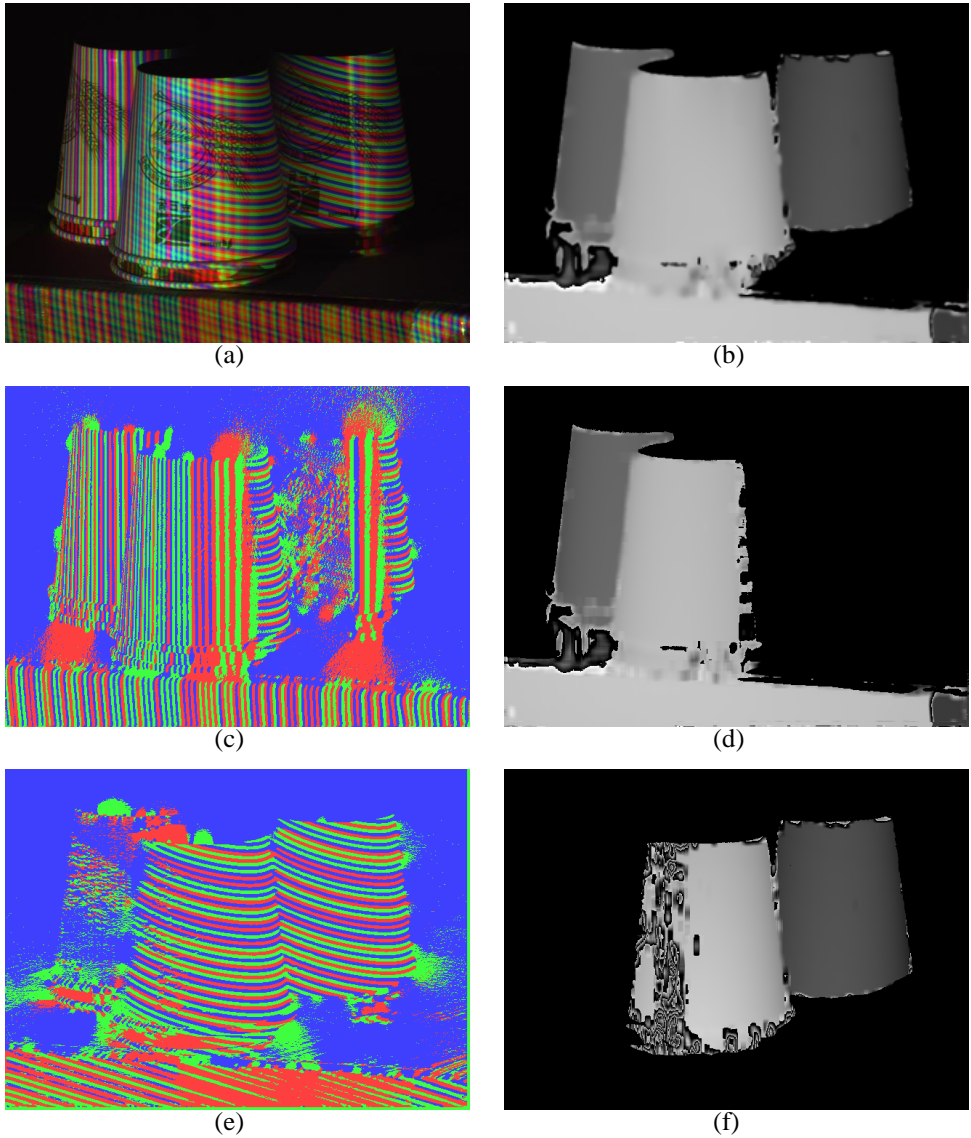

Figure 12. Results of paper cups. (a) The scene image of color-striped cups, (c,e) the stripe-segmented images of (a). The ranges (d) and (f) are obtained from the two stripe-segmented images, and they are merged into the one (b) which is much completer than the initial two ranges.

**Triple-pattern**: Figure 11 shows the results from a Venus face illuminated by three projectors. The three stripe-segmentation images are extracted from a single scene image. The three ranges are obtained from them, and they merge into one, which is meshed and rendered. Although only one camera image is used to estimate the surface geometry, the much complete shape is acquired by utilizing the three color-stripe patterns extracted from the single image.

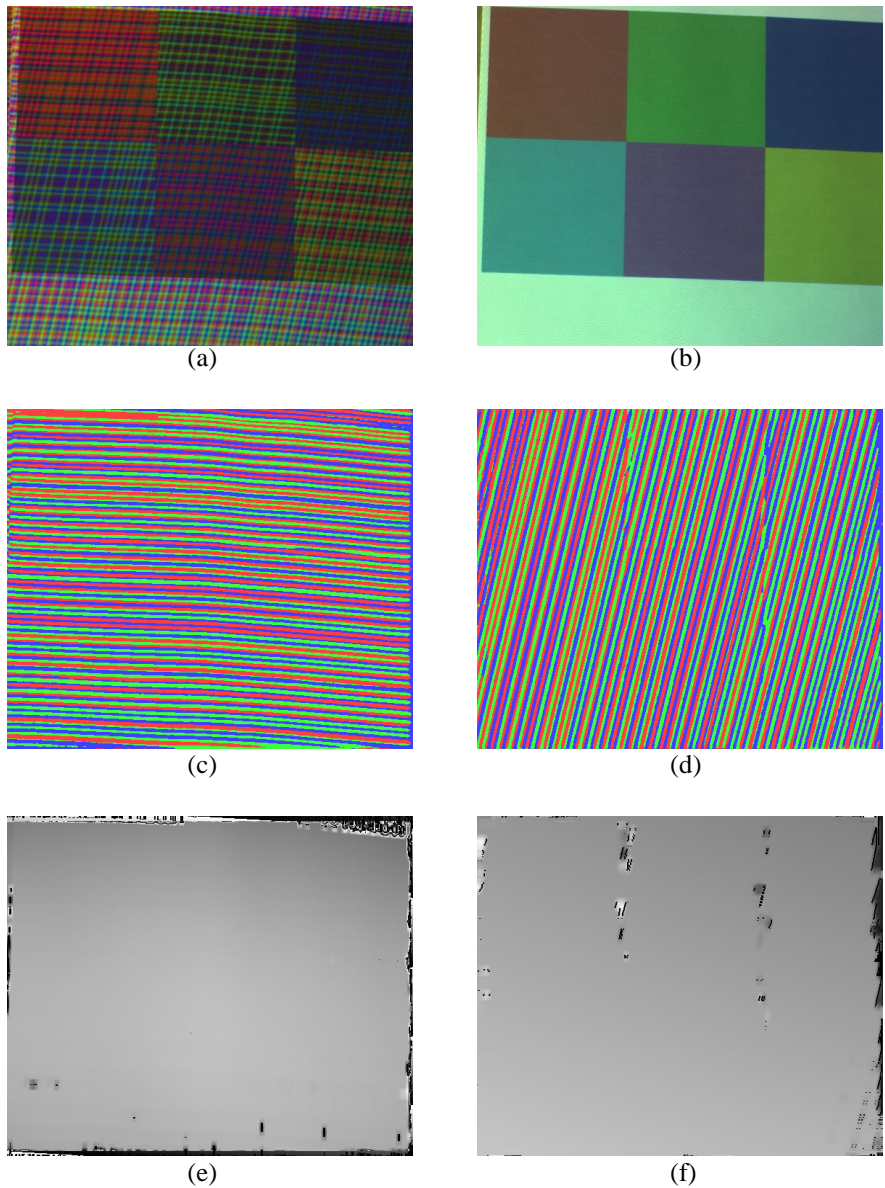

Figure 13. Experimental results of a panel with strong colors (RGBCMY). (a) The color panel with multi-projector structured-light patterns, (b) The color panel with white illumination, (c,d) the stripe-segmented images, and (e,f) the range results from the stripe-segmented images (c) and (d).

**Occlusion**: Figure 12 shows the results from paper cups with much extended occlusions. The two ranges (d) and (f) lack for either side, but a much completer range (b) is obtained by merging the two ranges. In merging the two ranges, we use a variant of median filter as follows. For each pixel location, valid range values in a 3×3 window are collected to be a set from each range image, so at most nine range values can be included in a set. Between the two sets, we select a set to compute the median as follows. If the difference between the numbers of components in the two sets is bigger than three, we select the set that has more components. Otherwise, we select the set that has a smaller variance (if the two variances are the same, we compute the median of the all components). There are some degenerate cases. If a set has only one component, the variance should be the maximum programmable value. If a set is empty, it should not be selected.

**Strong-colored object**: The presented method is also tested with a flat color panel that consists of RGBCMY squares. Figure 13 shows the experiments with the color panel. The image in (b) is not used in any process for obtaining range information. To diminish the influence of surface colors in the stripe segmentation, the intensity in the separated image is divided by the locally averaged color. Using this simple processing, the stripes are segmented accurately and the ranges are reliably obtained. Some errors exist around edges of surface color and near image boundary in (d), (e), and (f) because of the following reasons: (1) Pixels around edge between strong surface colors have gradients similar to the gradients of illumination pattern. (2) Gradients are not properly defined at boundary pixels, and thus some subpatterns might be falsely recognized around the image boundary even after multi-layer decoding [31].

## 5. Discussion and Implication

By using multiple projectors, multiple structured-light patterns can be projected into an object scene. Multi-projector lighting from the multiple directions increases the properly illuminated area, and decreases unwelcome effects of specularities and shading as well as shadows (most vision algorithms are seriously affected by them) by diffusing reflected rays under the existence of specularities and by counterbalancing shading effects. However, multi-projector structured-light vision has been prohibited by interference between multiple illumination patterns, which is considerably resolved by the proposed method.

SLV is one branch of the triangulation-based shape-measuring method that additionally includes its other branches, conventional passive stereo vision (PSV) and ASV. The most basic setup of PSV, SLV, and ASV consists of *camera + camera*, *projector + camera*, and *camera + projector + camera*, respectively. In general, SLV and ASV have better accuracy than PSV since they actively employ light patterns for simplifying the correspondence problem. Besides, SLV basically requires lower hardware cost compared to ASV, and is independent of the occlusions between viewing sensors. Based on our multi-projector color SLV, the required hardware, number of images and processing time for modeling a static/dynamic 3D object can be much more decreased than in conventional active/passive stereo methods. In addition, the number of range images obtainable from a single image corresponds to that of superposed light patterns in the image. Since multi-projector SLV allows superposing structured-light patterns and extracts the multiple patterns, it acquires a more amount of 3D data from multiple ranges, and reconstructs more complete geometry of an object. Table 2 compares relative amounts and completeness of 3D data obtainable by conventional SLV and multi-projector SLV. In the table, a 3D object is assumed topologically similar to a sphere, and the relative completeness represents the proportion of the solid angle of obtainable 3D data to that of the bounding sphere of the 3D object. The relative amount of data is the proportion of the amount of obtainable 3D data to that of the single-scan data of the 3D object under the assumption that the single scan is performed with similar resolutions in lighting and sensing. The relative amount of 3D data and relative completeness are computed based on five simple assumptions: (1) One, two, three, and four projector(s) properly illuminate 40%, 70%, 90%, and 100% of the object surface, respectively. (2) One, two, three, and four camera(s) properly capture 40%, 70%, 90%, and 100% of the object surface, respectively. (3) One, two, three, and four projector(s) properly illuminate 70%, 80%, 90%, and 100% of camera image(s), respectively. (4) One, two, three, and four camera(s) properly capture 70%, 80%, 90%, and 100% of projector projection(s), respectively. (5) Adding one more projector-camera triangulation results in 0.03% loss of 3D data amount per triangulation since the average error rate slightly increases with the number of triangulations.

Table 2
Relative Amounts and Completeness of 3D Data Obtainable by Conventional SLV and Multi-Projector SLV

| The number of projectors | The number of cameras | Relative amount of 3D data [%] | Relative completeness [%] |
|---|---|---|---|
| *1* | 1 | 28 | 28 |
| *1* | 2 | 54 | 32 |
| *2* | 1 | 54 | 32 |
| *2* | 2 | 102 | 56 |
| *3* | 1 | 79 | 36 |
| *3* | 2 | 144 | 63 |
| *3* | 3 | 198 | 81 |
| *4* | 4 | 284 | 100 |

**Comparison to Furukawa *et al*. [30]:** We originally presented most of our approach in a PhD thesis [32] about three years earlier than Furukawa *et al*. [30] which is the only closely related work (multi-projector SLV), currently. However, their work has the following differences compared to ours.

- Their work has no sense of generality for orientation cue, whereas ours does. They just employed the concept of two perpendicular directions (vertical and horizontal lines). On the other hand, we generalized orientation cue to the arbitrary angle between any two stripe patterns among an arbitrary number of superposed stripe patterns (practically up to three superposed patterns), and provided the analytic validation of the explicit derivative computations for separation of superposed patterns.

- For detecting lines of different projectors, they just rotated the image to be processed without any description on how the directions of patterns are determined. Moreover, this is similar to the way presented in our previous conference paper [26]. On the contrary, we provided the method of estimating stripe-encoding directions.

- The number of unique subpatterns is just eight for each projector pattern of theirs. It is much smaller than those in [12] (125) and ours (192), and thus has a strong disadvantage in that the projector-camera disparity should be assumed very small. This means practically a smaller triangulation angle which brings lower accuracy, and is improper for the capture of detailed surface geometry (compare their results to ours).

- They have not presented results of more challenging scenes such as the results of a real human face (Figure 9) and those of a strong-colored object (Figure 13) presented in this paper. By the way, the result of a hand in [30] actually seems to be a glove. Finger parts of a glove are substantially wider than fingers of a real hand, and thus attaining a good result for a glove is much easier than for a real hand (compare the results in [30] and [26]).

## 6. Conclusion and future work

Simultaneous use of multiple structured-light patterns for modeling a dynamic object has been a challenging problem since active light patterns from multiple sources can interfere with each other. After analyzing color phenomena in multi-projector light, we presented a method of using multiple structured-light projectors simultaneously for modeling a moving object in the temporal resolution of a video. A color-stripe permutation pattern suitable for range capture in a single video frame is employed, and a small number of the identical patterns are simultaneously projected onto a 3D object surface in different orientations. The presented technique extracts the original structured-light patterns based on the analysis of spatial color derivatives. The experimental results demonstrated the validity of our method. Our method is applicable to complete-3D modeling of dynamic objects.

Research on multi-projector SLV is now in its childhood. Superposition of illuminations from multiple light sources makes estimation of each illumination highly difficult by itself. In addition, variation of illumination patterns in multi-projector SLV will need more complex analyses and methods that are more delicate. In addition, increasing the number of superposed patterns in a certain area will raise the range errors, since the estimation of each illumination will be more sensitive to the limits in the dynamic range and SNR of cameras, which all should be investigated in the future work. When multiple cameras are employed together with single or multiple projectors for range imaging, additional range data can be obtained from the camera-camera correspondences based on ASV and PSV. In this regard, structured-light stereo [36] has been presented as a method of integrating SLV and ASV.

**Acknowledgements**

We thank the editor and reviewers for their valuable comments and suggestions. This research was supported by the Converging Research Center Program funded by the Ministry of Education, Science and Technology (2012K001342) and by the National Research Foundation of Korea (NRF) grant funded by the Korea government (MOE) (No. NRF-2012R1A1A2009461).